\documentclass{article}
\usepackage{cite}
\usepackage{amsmath,amssymb,amsfonts}
\usepackage{algorithmic}
\usepackage{graphicx}
\usepackage{textcomp}
\usepackage{url}
\usepackage{hyperref}
\usepackage{courier}

\title{Transformer Classification of Breast Lesions: The BreastDCEDL\_AMBL Benchmark Dataset and 0.92 AUC Baseline}

\author{Naomi Fridman$^{1,*}$ and Anat Goldstein$^{1}$\\
$^{1}$Department of Industrial Engineering, Ariel University, Ariel 40700, Israel\\
$^{*}$Corresponding author: naominoe.fridman@msmail.ariel.ac.il}

\date{}

\begin{document}

\maketitle

\begin{abstract}
Breast magnetic resonance imaging serves as a critical modality for cancer detection and treatment planning, yet its clinical utility is limited by poor specificity, resulting in high false-positive rates and unnecessary biopsies. This study presents a transformer-based framework for automated classification of breast lesions in dynamic contrast-enhanced MRI, addressing the fundamental challenge of differentiating benign from malignant findings. We implemented a SegFormer architecture that achieved 0.92 AUC for lesion-level classification, demonstrating 100\% sensitivity and 67\% specificity at the patient level—potentially eliminating one-third of unnecessary biopsies without missing malignancies. The model quantifies malignant pixel distribution through semantic segmentation, generating interpretable spatial predictions that support clinical decision-making. To establish reproducible benchmarks, we curated BreastDCEDL\_AMBL by transforming The Cancer Imaging Archive's AMBL collection into a standardized deep learning dataset containing 88 patients with 133 annotated lesions (89 benign, 44 malignant). This resource addresses a critical infrastructure gap, as existing public datasets lack benign lesion annotations, preventing benign-malignant classification research. Training utilized an expanded cohort exceeding 1,200 patients through integration with BreastDCEDL datasets, validating transfer learning approaches despite primary-tumor-only annotations. Public release of the dataset, models, and evaluation protocols provides the first standardized benchmark for DCE-MRI lesion classification, enabling methodological advancement toward clinical deployment.
\end{abstract}

\noindent\textbf{Keywords:} Breast MRI, DCE-MRI, lesion classification, public dataset, SegFormer

\section{Introduction}
\label{sec:introduction}

Breast cancer remains the most diagnosed malignancy among women worldwide, with dynamic contrast-enhanced magnetic resonance imaging (DCE-MRI) serving as an essential tool for detection, diagnosis, and treatment planning. While DCE-MRI achieves unparalleled sensitivity exceeding 95\% for cancer detection, its clinical impact is constrained by poor specificity, resulting in a substantial burden of false-positive findings. This specificity challenge manifests most acutely in clinical practice, where approximately 65\% of MRI-guided breast biopsies reveal benign pathology, subjecting patients to unnecessary invasive procedures, psychological distress, and significant healthcare costs. The development of automated classification systems capable of accurately distinguishing benign from malignant lesions represents a critical need in breast imaging, with the potential to preserve MRI's exceptional sensitivity while reducing false-positive rates. This work presents a transformer-based deep learning approach and accompanying public dataset that addresses this fundamental challenge, demonstrating that high-performance lesion classification is achievable through the integration of advanced neural architectures with comprehensive training data.

\subsection{Scientific Background}

Breast MRI has evolved into a key diagnostic modality for breast cancer detection and characterization, with primary indications including evaluation of lesions detected by mammography/ultrasound, assessment of disease extent in newly diagnosed patients, monitoring therapeutic response, and screening high-risk populations~\cite{sardanelli2010eusoma_1}. Compared to mammography and ultrasound, MRI provides superior sensitivity for breast cancer detection, enabling identification of malignancies occult on physical examination and conventional imaging~\cite{kuhl2005surveillance_2}.

Standard breast MRI protocols rely on Dynamic Contrast-Enhanced (DCE) sequences that track tissue enhancement following intravenous gadolinium administration. These T1-weighted sequences are acquired before and at multiple timepoints (typically 4--8) after contrast injection~\cite{kuhl2007currentI_3}, capturing both the initial enhancement (wash-in) and dynamic patterns over subsequent minutes. Three characteristic kinetic patterns emerge: rapid wash-in followed by washout (suggestive of malignancy), plateau enhancement, or persistent gradual uptake (typically benign)~\cite{wang2009kinetic_4}. Current DCE-MRI analysis builds upon the kinetic modeling framework established by Degani et al.~\cite{degani1997mapping_5}, which quantifies these temporal enhancement dynamics.

Despite exceptional sensitivity, breast DCE-MRI suffers from limited specificity. Early studies consistently reported false-positive rates approaching 60\% for MRI-guided biopsies due to benign enhancing lesions and non-mass-like enhancement that mimic malignancy~\cite{kuhl2007currentII_6,kriege2006factors_7,baltzer2010birads_8}. More recent systematic reviews and large institutional cohorts have found that the rate of benign findings is even higher, averaging around 65\% (95\% CI: 59--73\%) of all MRI-guided breast biopsies~\cite{ozcan2022benchmark_9,motanagh2023trends_10}. The particular combinations of morphologic and kinetic features that optimally discriminate benign from malignant lesions remain undefined, necessitating improved computational approaches for personalized diagnosis and biopsy guidance.

\subsection{Previous Work and Current Limitations}

Deep learning, particularly convolutional neural networks (CNNs), has been extensively investigated for breast cancer classification in DCE-MRI~\cite{zhu2023cnn_11}. These networks automatically learn discriminative features from imaging data~\cite{gomezflores2025ranklet_12}, eliminating manual feature engineering. However, their millions of trainable parameters require substantial annotated data, leading most researchers to employ transfer learning strategies.

Recent advances demonstrate promising results, as summarized in Table~\ref{tab:recent_studies}. Nascimento et al.~\cite{nascimento2025foundation_13} developed a Medical Slice Transformer (MST) combined with Kolmogorov-Arnold Networks, achieving AUC 0.80 with attention-based interpretability on 232 private cases. Gómez-Flores et al.~\cite{gomezflores2025ranklet_12} employed 3D ranklet-based texture analysis on 97 DCE-MRI exams, achieving AUC 0.89. Wang et al.~\cite{wang2023mobilenet_14} reported AUC up to 0.93 using feature-level Maximum Intensity Projections, while Li et al.~\cite{kinetic2025arxiv_15} achieved AUC 0.94 combining radiomics with kinetic features. Gullo et al.~\cite{gullo2025mlmr_16} matched expert radiologist performance (AUC $\geq$0.91) for non-mass enhancement classification.

\begin{table}[ht]
\centering
\caption{Comparative Analysis of Recent Deep Learning Approaches for Breast Lesion Classification in DCE-MRI}
\label{tab:recent_studies}
\begin{tabular}{|p{1.8cm}|p{1.5cm}|p{7.7cm}|}
\hline
\textbf{Study} & \textbf{Dataset} & \textbf{Methodology and Performance} \\
\hline
Nascimento et al. (2025)~\cite{nascimento2025foundation_13} & 3,505\newline exams\newline propriety & Medical Slice Transformer with attention mechanisms for interpretability. Achieved AUC of 0.80, outperforming Vision Transformer baselines. \\
\hline
Gómez-Flores et al. (2025)~\cite{gomezflores2025ranklet_12} & 97\newline exams\newline propriety & 3D ranklet transform for texture-based feature extraction. Demonstrated AUC of 0.89 for benign-malignant discrimination. \\
\hline
Wang et al. (2023)~\cite{wang2023mobilenet_14} & 687 \newline exams\newline propriety & Fine-tuned MobileNet with feature-level Maximum Intensity Projections. Reported AUC of 0.93 through volumetric-temporal integration. \\
\hline
Li et al. (2024)~\cite{kinetic2025arxiv_15} & 554 \newline exams\newline propriety & Hybrid framework integrating radiomics with kinetic curve analysis. Achieved AUC of 0.94 through validated feature fusion. \\
\hline
Gullo et al. (2025)~\cite{gullo2025mlmr_16} & 573 \newline NME \newline lesions\newline propriety  & ResNet architecture for non-mass enhancement classification. Achieved AUC $\geq$ 0.91, matching expert radiologist performance. \\
\hline
\end{tabular}
\end{table}

Critically, all state-of-the-art research relies on private datasets, severely limiting reproducibility and clinical translation. Available public datasets have fundamental limitations: it contains only primary malignant tumor segmentations per standard medical protocols, preventing benign/malignant classification research~\cite{khaled2022unet_17}. BreastDM provides only partial slices reduced to 8-bit JPEGs, sacrificing critical intensity information~\cite{zhao2023breastdm_18}.

The generalization challenge is well-documented. Horvat et al.~\cite{horvat2024radiomics_19} concluded that ``radiomics features often demonstrate high performance only when applied to data similar to that on which they were originally trained'', with differences in acquisition protocols and scanner types leading to non-reproducible results.

\subsection{Contribution}

We address these critical gaps through two primary contributions. First, we curated BreastDCEDL\_AMBL from the publicly available AMBL dataset hosted on The Cancer Imaging Archive (TCIA)  ~\cite{daciels2025ambl_20}---the only publicly available DCE-MRI collection containing comprehensive lesion segmentations for both benign and malignant lesions with preserved raw data integrity. Our dataset comprises 88 patients with complete T1-weighted sequences (one pre-contrast, four post-contrast) and 133 annotated lesions (89 benign, 44 malignant), converted to 3D NIfTI format preserving original signal values.

Second, we developed a SegFormer-based classification pipeline trained on an expanded cohort incorporating BreastDCEDL ~\cite{fridman2025breastdcedl_21}, totaling over 1,200 patients. Our approach achieves AUC 0.92 on independent test data with 100\% sensitivity at the patient level through optimized decision thresholds. By releasing this curated dataset, trained models, and standardized evaluation protocols, we establish a reproducible benchmark enabling fair comparison of future methods and advancing the field toward clinically deployable solutions.

\section{Methods}

\subsection{Dataset Curation and Preprocessing}

We curated a subset of 88 patients from the AMBL dataset available in The Cancer Imaging Archive (TCIA)~\cite{daciels2025ambl_20}, a single-institutional retrospective collection of 632 breast DCE-MRI examinations acquired on a 1.5T MR system between 2018--2021. The selection criteria required complete T1-weighted DCE sequences comprising five temporal acquisitions (one pre-contrast and four post-contrast phases) with comprehensive lesion segmentation annotations. Importantly, the term ``lesion'' in this dataset encompasses both mass and non-mass enhancement (NME) types, consistent with BI-RADS definitions. Lesion annotation was performed manually and supplemented by automated thresholding (pixels exceeding intensity 75 in subtraction images).

For deep learning compatibility, raw DICOM~\cite{iso2025dicom_22} slice files were collected and organized to their 3D spatial structure, converted to standardized NIfTI format~\cite{nifti2001standard_23} while preserving signal intensities. Each patient's DCE acquisitions were reorganized into separate 3D NIfTI volumes. Segmentation annotations were structured into two NIfTI masks per patient: (1) a comprehensive lesion mask (\texttt{<patient\_id>\_sus}) with unique integer labels for every lesion (benign or malignant), and (2) a malignant-specific mask (\texttt{<patient\_id>\_tum}) that includes only malignant tumor segmentations. This dual-mask format supports spatial detection and subsequent classification, while maintaining lesion-level granularity. BreastDCEDL\_AMBL matches the organizational structure of BreastDCEDL and is publicly accessible on Zenodo~\cite{fridman2025amblzenodo_24}, with code and guides at \url{https://github.com/naomifridman/BreastDCEDL_AMBL}.

BreastDCEDL\_AMBL dataset comprises 133 annotated lesions (89 benign, 44 malignant) distributed among 88 patients. The cohort contains 31 patients presenting malignant tumors, 42 with benign lesions alone, and 15 with both lesion types. Stratified random sampling produced training, validation, and test sets with proportional malignant/benign cases and HER2 receptor status (available for 47 patients). These splits, summarized in Table~\ref{tab:dataset_stats}, ensure reproducibility and allow for stratified biomarker analysis.

\begin{table}[t]
\centering
\caption{Dataset Summary and Partition Statistics}
\label{tab:dataset_stats}
\setlength{\tabcolsep}{3pt}
\begin{tabular}{|l|c|c|c|c|}
\hline
\textbf{Characteristic} & \textbf{Training} & \textbf{Validation} & \textbf{Test} & \textbf{Total} \\
& \textbf{(n=58)} & \textbf{(n=15)} & \textbf{(n=15)} & \textbf{(n=88)} \\
\hline
Patients with malignancy & 20 & 5 & 6 & 31 \\
Patients with benign findings & 28 & 7 & 7 & 42 \\
Malignant lesions & 29 & 7 & 8 & 44 \\
Benign lesions & 59 & 15 & 15 & 89 \\
Total lesions & 88 & 22 & 23 & 133 \\
HER2-positive cases & 14 & 4 & 4 & 22 \\
HER2-negative cases & 17 & 4 & 4 & 25 \\
\hline
\end{tabular}
\end{table}

\begin{figure*}[t]
\centering
\includegraphics[width=\textwidth]{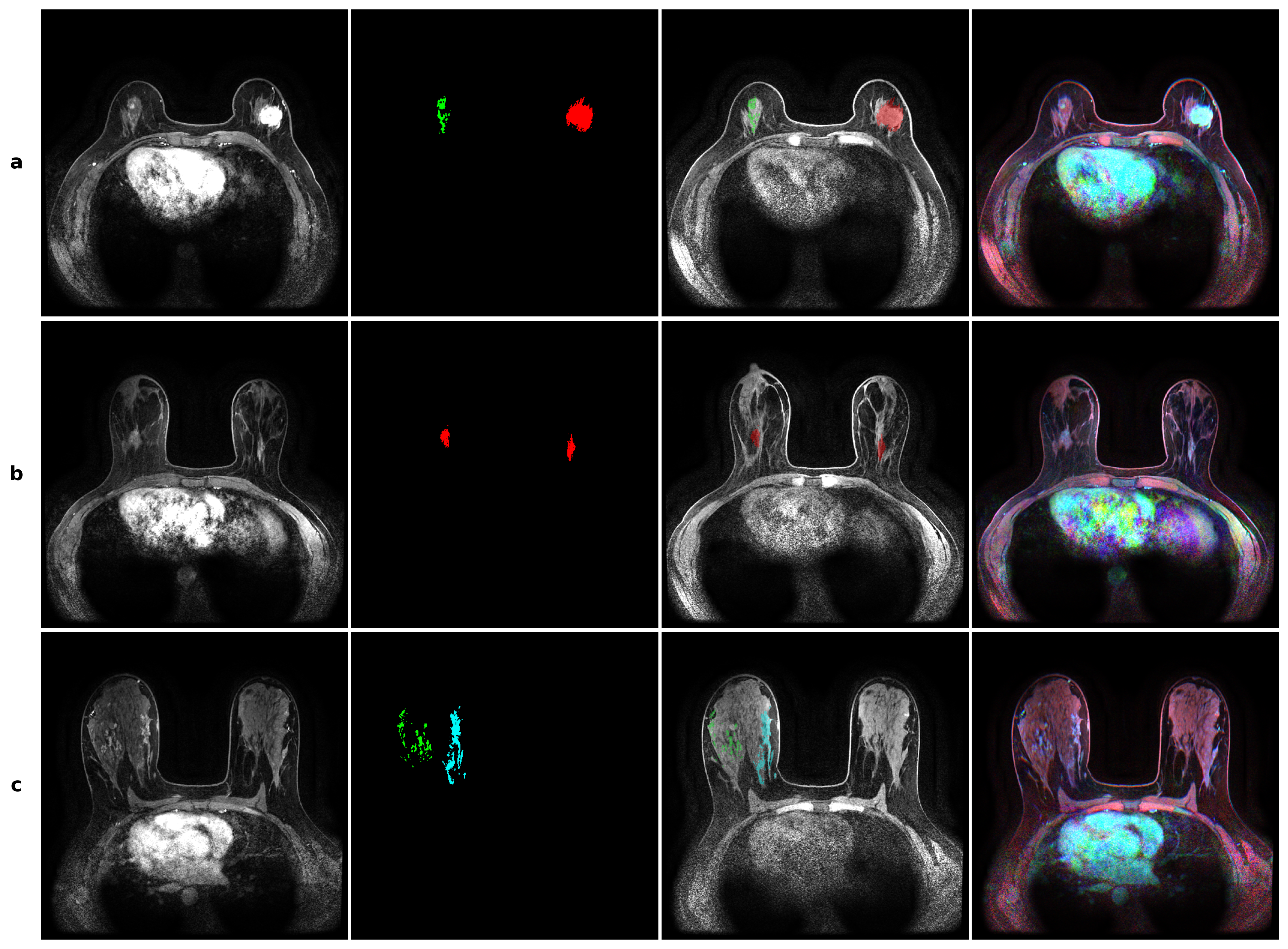}

\caption{Representative lesion segmentation examples from the AMBL dataset showing three patients (rows a--c). Each row displays: peak enhancement phase (column 1), color-coded segmentation with benign (green/blue) and malignant (red) lesions (column 2), segmentation overlay on peak enhancement (column 3), and RGB fusion showing temporal enhancement kinetics (column 4). Patient (a) shows bilateral lesions with one benign and one malignant. Patient (b) demonstrates bilateral malignant lesions. Patient (c) presents bilateral benign lesions. The RGB fusion highlights contrast uptake and washout patterns for lesion characterization.}
\label{fig:segmentation_examples}
\end{figure*}

\subsection{Training Data Expansion Using BreastDCEDL Datasets}

Given the relatively small size of our curated BreastDCEDL\_AMBL dataset (88 patients), we expanded our training cohort by incorporating additional cases from BreastDCEDL-ISPY1 and BreastDCEDL-ISPY2, which provide 163 and 982 patients respectively with complete DCE-MRI sequences and tumor segmentations~\cite{fridman2025breastdcedl_21,fridman2025dcedlzenodo_25}. These datasets, derived from The Cancer Imaging Archive (TCIA) and processed into standardized NIfTI format as described in~\cite{fridman2025dcedlzenodo_25}, contain comprehensive primary tumor annotations generated through hybrid manual and computer-aided detection methods by experienced radiologists.

\begin{figure*}[t]
\centering
\includegraphics[width=\textwidth]{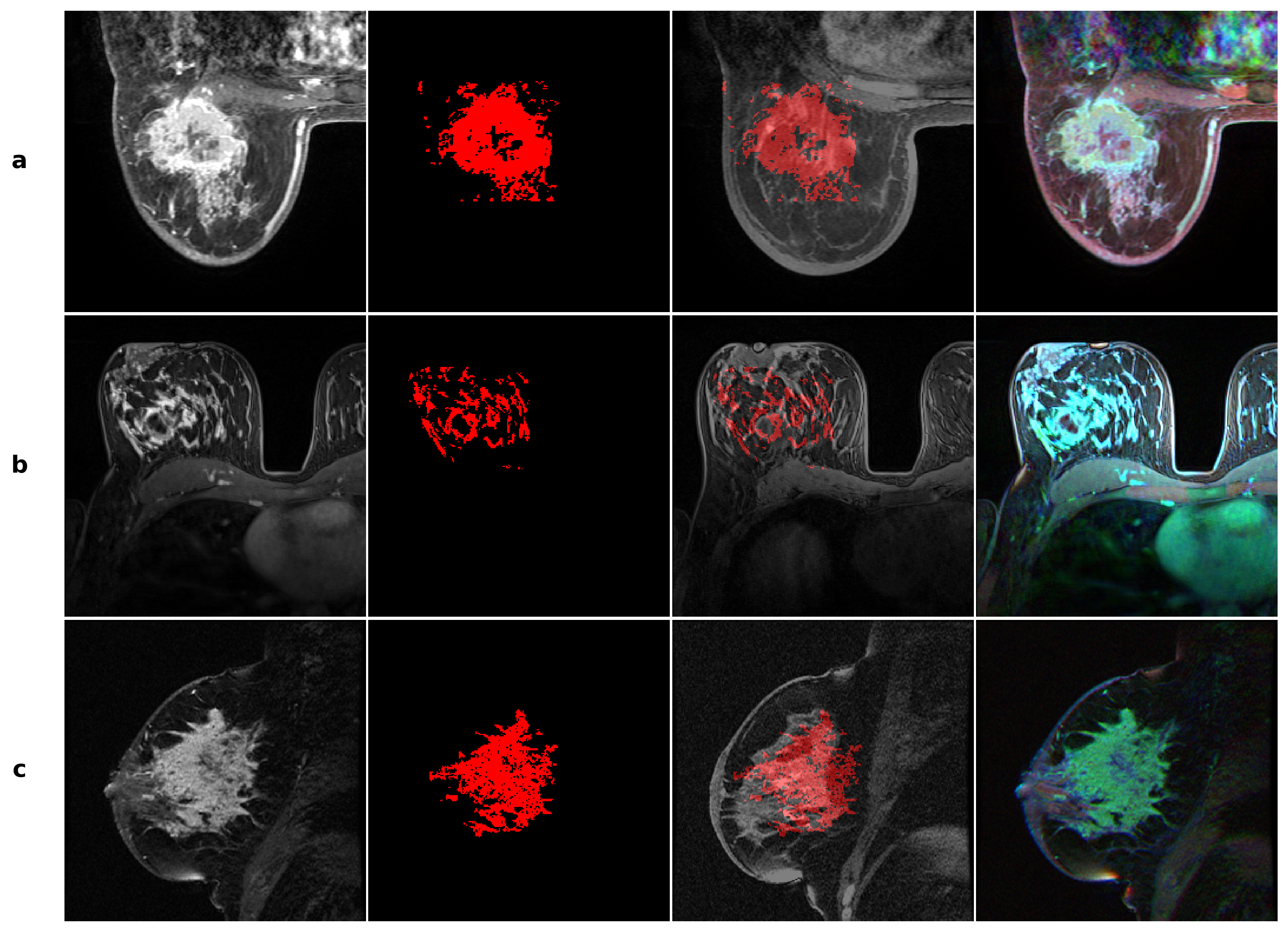}
\caption{Representative lesion segmentation examples from BreastDCEDL datasets. Three patient cases are shown: rows (a) and (b) from BreastDCEDL\_ISPY2, and row (c) from BreastDCEDL\_ISPY1. Column 1 displays the peak enhancement phase following contrast administration. Column 2 shows the binary tumor segmentation mask. Column 3 presents the segmentation overlay on the peak enhancement image. Column 4 illustrates the RGB fusion combining temporal DCE-MRI phases, where pre-contrast, early post-contrast, and late post-contrast acquisitions are mapped to red, green, and blue channels respectively, enabling visualization of enhancement dynamics.}
\label{fig:ispy_examples}
\end{figure*}

As illustrated in Figure~\ref{fig:ispy_examples}, the ISPY datasets demonstrate diverse tumor presentations across different DCE-MRI phases, with row (a) showing an ISPY2 case and rows (b--c) depicting ISPY1 examples. Each case displays the peak enhancement phase (column 1), tumor segmentation mask (column 2), segmentation overlay (column 3), and RGB fusion of temporal phases (column 4). Importantly, while breast MRI scans often contain multiple lesions, the ISPY datasets provide segmentations only for the primary tumor, leaving any additional benign or malignant lesions unannotated. 

To adapt these data for our segmentation task, we implemented a targeted extraction strategy: $256 \times 256$ pixel patches were cropped around the annotated primary tumors. This cropping served dual purposes: (1) it created appropriately sized inputs for our segmentation network, and (2) it potentially captured unannotated secondary lesions present in the surrounding tissue, though these would be treated as background in the training labels. This approach was particularly suitable for the ISPY cohorts, which predominantly feature larger primary tumors, ensuring that the cropped regions maintained the primary lesion as the central feature while including perilesional tissue context. This strategy expanded our training set from 88 to over 1,200 patients, providing the data volume necessary for robust deep learning model training while maintaining compatibility with our segmentation framework.

\subsection{Deep Learning Architecture Selection}

For breast lesion segmentation, we evaluated multiple state-of-the-art architectures before selecting SegFormer~\cite{xie2021segformer_26} as our backbone model. While convolutional neural networks have traditionally dominated medical image segmentation, their limited receptive fields struggle to capture the long-range dependencies essential for understanding global tissue context in DCE-MRI. Among transformer-based alternatives---including Vision Transformer (ViT)~\cite{dosovitskiy2020imagewords_27}, SETR~\cite{zheng2020seq2seq_28}, Swin Transformer~\cite{liu2021swin_29}, and TransUNet~\cite{chen2024transunet_30}---SegFormer emerged as optimal for addressing the specific challenges of breast MRI analysis.

\begin{figure*}[t]
\centering
\includegraphics[width=\textwidth]{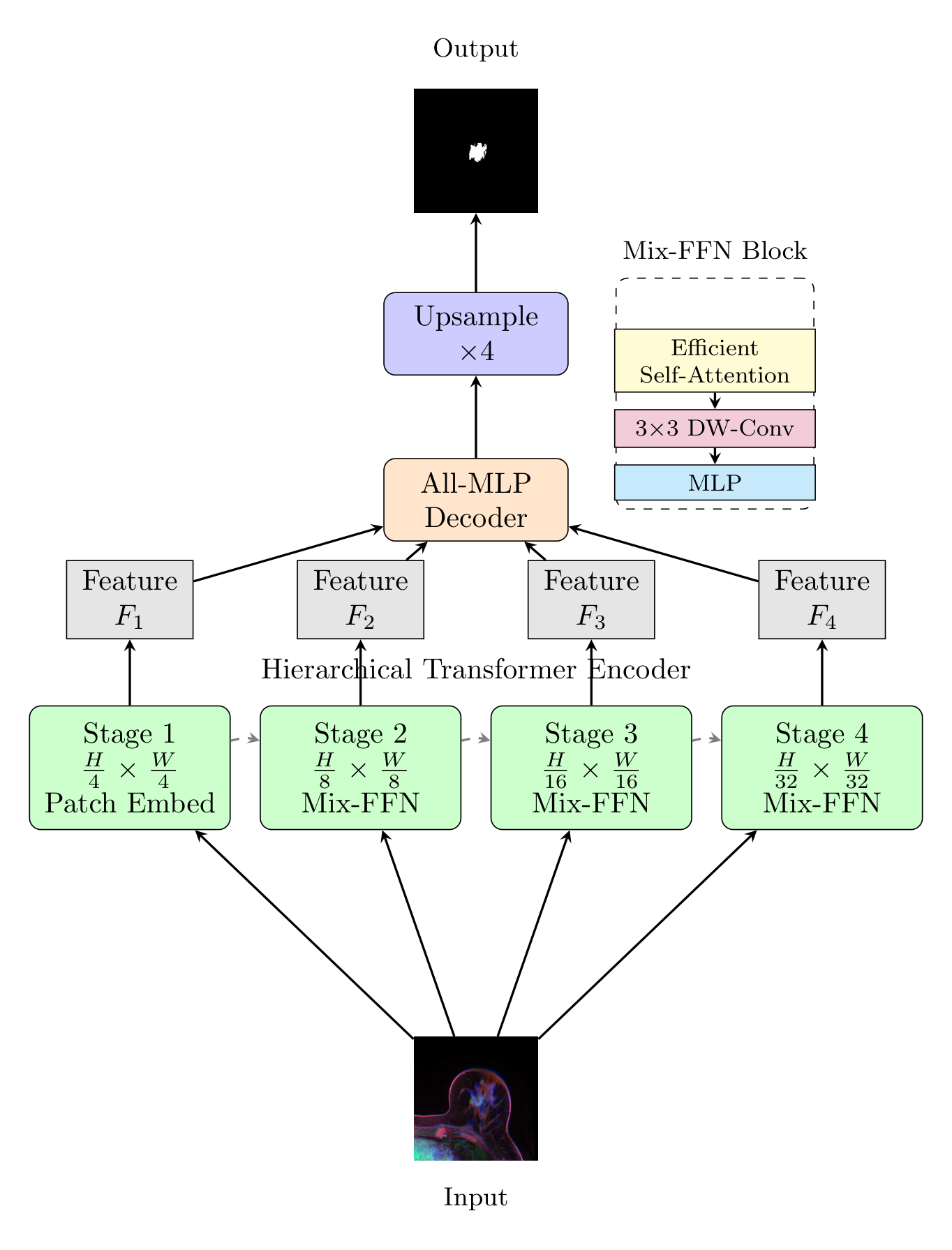}
\caption{SegFormer architecture for breast lesion segmentation. The model consists of a hierarchical transformer encoder with four stages that progressively reduce spatial resolution from $H/4 \times W/4$ to $H/32 \times W/32$ while extracting multi-scale features ($F_1$--$F_4$). Stage 1 employs patch embedding for initial feature extraction, while Stages 2--4 utilize Mix-FFN blocks that combine efficient self-attention with $3 \times 3$ depthwise convolutions for implicit positional encoding. The lightweight All-MLP decoder aggregates these multi-scale features before upsampling by a factor of 4 to produce the final segmentation mask at original resolution.}
\label{fig:segformer_architecture}
\end{figure*}

Breast DCE-MRI data presents unique computational challenges: volumes acquired across different scanners and protocols exhibit substantial variability in spatial resolution ($256 \times 256$ to $1024 \times 1024$ pixels), slice thickness, and field of view. As illustrated in Figure~\ref{fig:segformer_architecture}, SegFormer's hierarchical architecture addresses these challenges through progressive feature extraction across four stages, reducing spatial resolution from $H/4 \times W/4$ to $H/32 \times W/32$ while maintaining computational efficiency. The overlapping patch embedding in Stage 1 preserves local continuity at lesion boundaries---crucial for distinguishing irregular malignant margins from smooth benign contours---unlike ViT's non-overlapping patches that may fragment boundary information~\cite{dosovitskiy2020imagewords_27}. The Mix-FFN blocks incorporate $3 \times 3$ depthwise convolutions providing implicit positional encoding, eliminating the need for fixed positional embeddings required by ViT and enabling processing at native resolutions without interpolation artifacts.

The lightweight All-MLP decoder aggregates multi-scale features ($F_1$--$F_4$) from all encoder stages, significantly reducing parameters compared to TransUNet's CNN decoder~\cite{chen2024transunet_30} or SETR's complex upsampling modules~\cite{zheng2020seq2seq_28}, while maintaining segmentation accuracy. This efficiency enables rapid inference across entire MRI volumes---essential for clinical deployment. The hierarchical design captures both fine-grained texture patterns and global morphological characteristics that differentiate benign from malignant lesions, combining the efficiency of CNNs with the global context modeling of transformers, making SegFormer uniquely suited for our breast lesion classification task.

\subsection{Data Preprocessing}

For input preparation, we created RGB fusion images by combining three temporal DCE-MRI acquisitions: pre-contrast, first post-contrast, and the final sequence (typically the fifth acquisition). Prior to fusion, each 2D slice underwent MinMax normalization independently, scaling intensity values to the $[0,1]$ range using the transformation $x' = (x - \min(x))/(\max(x) - \min(x))$, ensuring consistent signal representation across different acquisition protocols and scanner variations. The normalized pre-contrast, early post-contrast, and late post-contrast images were then mapped to the red, green, and blue channels respectively, creating pseudo-color representations that encode temporal enhancement dynamics.

Given the spatial heterogeneity across datasets---with ISPY trials exhibiting variable voxel dimensions, slice thickness, and in-plane resolution---we deliberately avoided spatial resampling to preserve the native image characteristics and prevent interpolation artifacts. Instead, we adopted a consistent cropping strategy: $256 \times 256$ pixel patches were extracted around each lesion center from the original $512 \times 512$ images. This approach was necessitated by the annotation limitations in ISPY1 and ISPY2 datasets, which provide segmentations only for the primary tumor despite the potential presence of additional unannotated benign and malignant lesions. Since ISPY cohorts predominantly contain larger tumors (mean diameter $>$2\,cm), the $256 \times 256$ cropping window typically captured the entire primary lesion while including surrounding tissue. We applied identical cropping to the BreastDCEDL\_AMBL dataset to maintain consistency across all training and inference data.

While SegFormer's architecture supports variable input sizes, preliminary experiments using full-resolution images yielded suboptimal results, likely due to the substantial size disparity between the typically larger ISPY tumors and the more diverse lesion sizes in the AMBL cohort. The standardized $256 \times 256$ input size provided optimal balance between computational efficiency and preservation of diagnostic features necessary for accurate lesion classification.

\subsection{Model Training and Optimization}

For each lesion across the ISPY1, ISPY2, and AMBL datasets, we extracted $256 \times 256$ pixel patches with 2-pixel padding around the lesion boundary, as perilesional features---particularly the tumor edge and adjacent tissue---have been identified as important predictors of malignancy in breast MRI analysis~\cite{antropova2018mipcnn_31,li2016radiomics_32}. Prior to RGB fusion, each cropped slice underwent MinMax normalization. Additional normalization strategies were employed as augmentation techniques, including clipping at the 0.01 quantile before normalization and applying MinMax normalization across all slices post-fusion.

To optimize our segmentation-based classification approach, we employed a hybrid loss function combining Binary Cross-Entropy (BCE) with Dice loss:
\begin{equation}
\mathcal{L}_{\text{combined}} = \mathcal{L}_{\text{BCE}} + \mathcal{L}_{\text{Dice}}
\end{equation}
where the BCE loss is defined as:
\begin{equation}
\mathcal{L}_{\text{BCE}} = -\frac{1}{N} \sum_{i=1}^{N} [t_i \log(p_i) + (1-t_i)\log(1-p_i)]
\end{equation}
and the Dice loss is:
\begin{equation}
\mathcal{L}_{\text{Dice}} = 1 - \frac{2\sum_{i} p_i t_i + \varepsilon}{\sum_{i} p_i + \sum_{i} t_i + \varepsilon}
\end{equation}
Here, $p_i \in [0,1]$ represents the predicted probability and $t_i \in \{0,1\}$ the ground truth label for pixel $i$, with $\varepsilon = 10^{-6}$ preventing division by zero.

This dual-objective formulation addresses the unique challenges of lesion segmentation: BCE ensures accurate pixel-level classification with strong gradients for decision boundaries, while Dice loss mitigates class imbalance by focusing on overlap ratios---crucial when lesions occupy small image regions. The combined loss directly supports our classification pipeline where lesions are deemed malignant based on the proportion of positive pixels within the segmented region. We empirically evaluated thresholds of 30\% and 50\%, finding that the 30\% threshold yielded superior performance on both training and validation sets.

During training, we employed comprehensive data augmentation to improve model robustness (detailed in our previous work~\cite{fridman2025breastdcedl_21}). Our pipeline included spatial augmentations (random cropping at scales $1.25\times$, $1.5\times$, $1.75\times$, or $2\times$ before resizing to $256 \times 256$, horizontal/vertical flipping, and $90^{\circ}$/$180^{\circ}$/$270^{\circ}$ rotations) and photometric augmentations (brightness adjustment $[0.7, 1.3]$, contrast modification $[0.8, 1.2]$, and Gaussian noise with $\sigma = 5$), all applied stochastically with 50\% probability. Transformations were applied consistently to both images and masks using appropriate interpolation (bilinear for images, nearest-neighbor for masks).

\begin{figure*}[t]
\centering
\includegraphics[width=\textwidth]{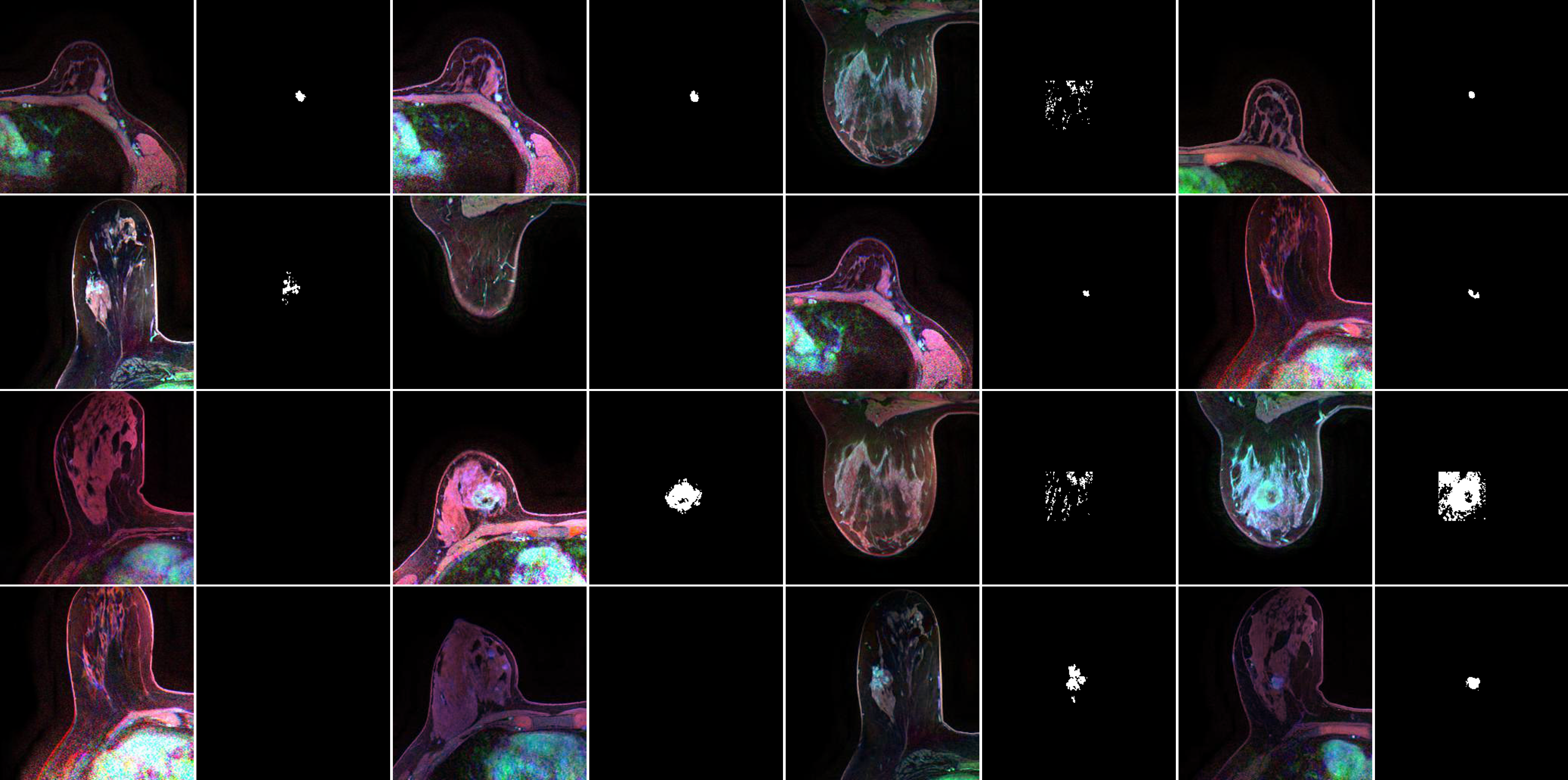}
\caption{Representative examples from a validation batch used during training, showing RGB-fused DCE-MRI slices (left) and their corresponding segmentation masks (right). Approximately half of the masks are empty (all-zero), corresponding to benign lesions in the BreastDCEDL\_AMBL dataset where both benign and malignant lesions are fully annotated.}
\label{fig:validation_examples}
\end{figure*}

\begin{figure*}[t]
\centering
\includegraphics[width=\textwidth]{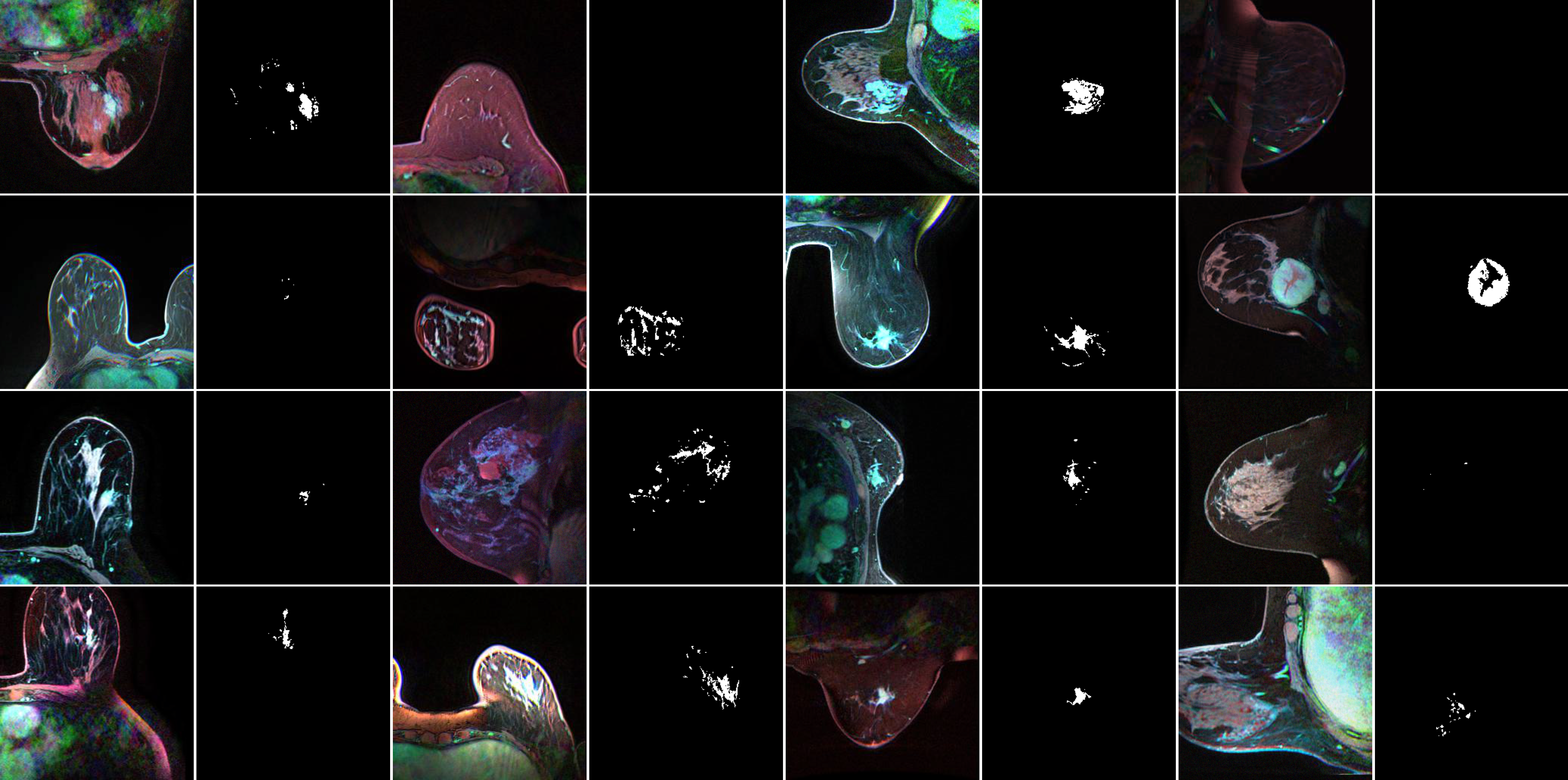}
\caption{Examples from a training batch combining all three datasets (BreastDCEDL\_ISPY1, BreastDCEDL\_ISPY2, and BreastDCEDL\_AMBL) with full augmentation applied. The images demonstrate varied lesion appearances and the effects of spatial and photometric augmentations, including rotation, scaling, and intensity adjustments.}
\label{fig:training_batch}
\end{figure*}

\begin{figure}[t]
\centering
\includegraphics[width=\columnwidth]{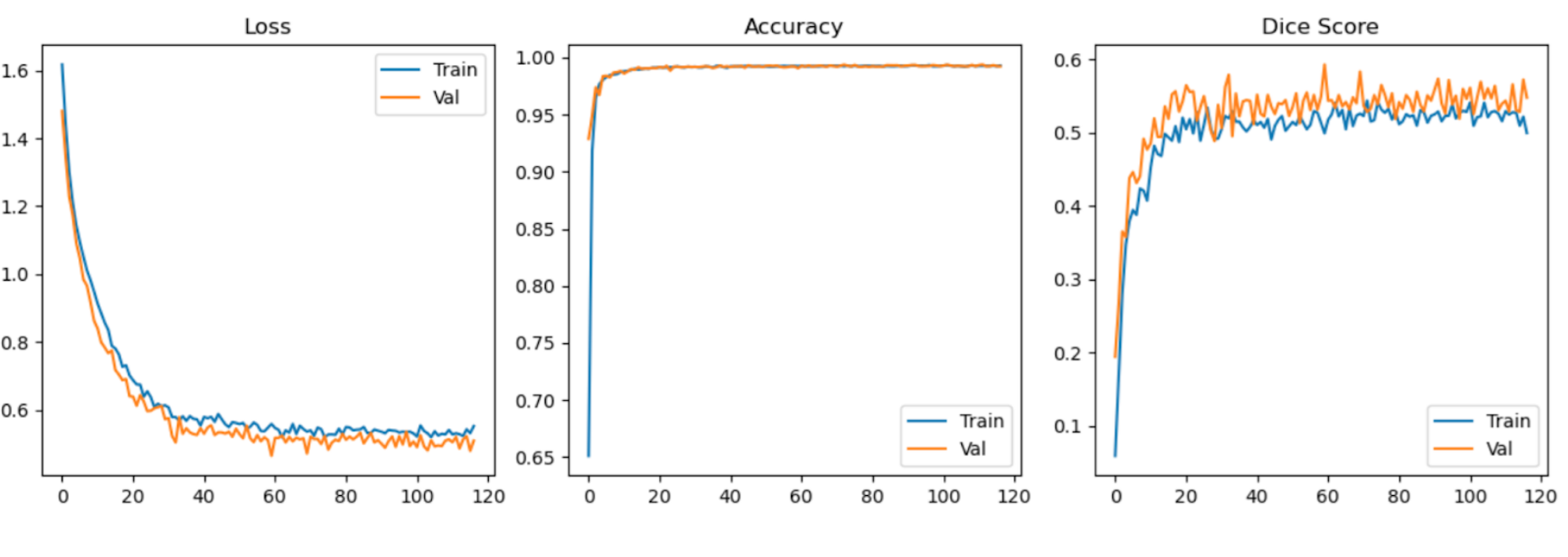}
\caption{Training metrics over 120 epochs. (a) Loss curves converge at approximately 0.5 by epoch 40. (b) Accuracy reaches 99\% for both training and validation sets. (c) Dice scores stabilize at approximately 0.5. Close alignment between training (blue) and validation (orange) metrics indicates robust generalization without overfitting.}
\label{fig:training_curves}
\end{figure}

Training was conducted for 120 epochs using cosine learning rate decay from $10^{-5}$ to $10^{-6}$, with epoch 59 selected based on optimal validation performance. Figure~\ref{fig:validation_examples} presents validation examples from BreastDCEDL\_AMBL, notably showing many empty (all-zero) segmentation masks corresponding to benign lesions, which are fully annotated in this dataset. In contrast, Figure~\ref{fig:training_batch}'s training batch contains fewer empty masks since BreastDCEDL\_ISPY1 and BreastDCEDL\_ISPY2 provide only malignant tumor segmentations. Figure~\ref{fig:training_curves} demonstrates stable training convergence with loss stabilizing around epoch 40, accuracy reaching $\sim$99\%, and Dice scores plateauing at $\sim$0.5, with validation metrics closely tracking training performance.


\subsection{Inference and Classification Pipeline}

\begin{figure*}[t]
\centering
\includegraphics[width=\textwidth]{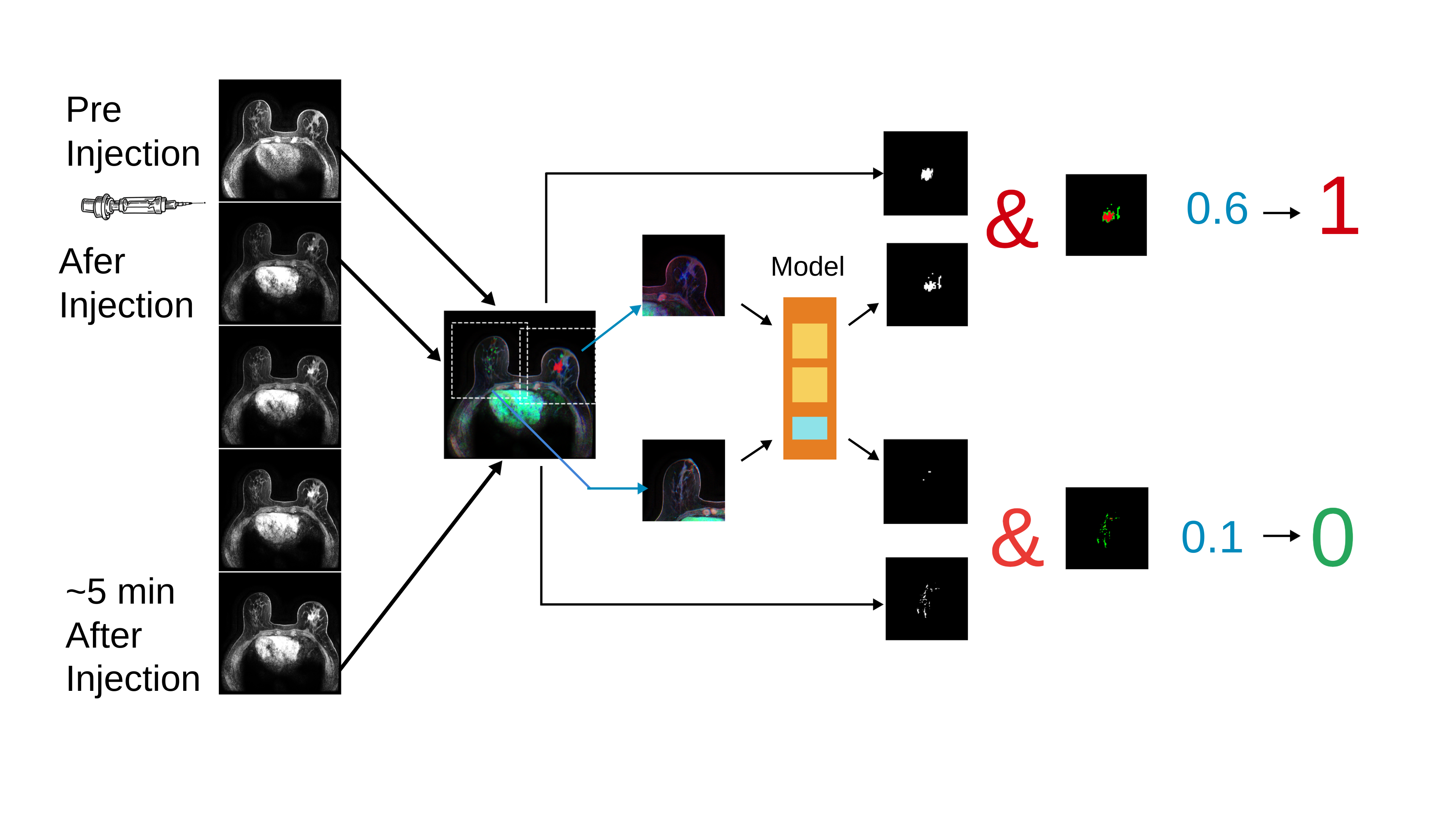}
\caption{Inference pipeline for multi-lesion breast cancer classification. DCE-MRI sequences (pre-contrast, post-contrast, and approximately 5 minutes post-injection) are RGB-fused to highlight enhancement patterns. Each identified lesion is cropped to $256 \times 256$ patches and processed through the SegFormer model (center, orange/yellow blocks). The model outputs segmentation masks (white indicates predicted malignant tissue) for each lesion. Malignancy scores are calculated as the ratio of malignant pixels to total lesion pixels (0.6 for upper lesion, 0.1 for lower lesion).}
\label{fig:inference_pipeline}
\end{figure*}

During inference, our trained SegFormer model processes new patients to classify individual breast lesions as benign or malignant. The pipeline handles multiple lesions per patient, as clinical examinations frequently present with more than one suspicious finding requiring evaluation. Figure~\ref{fig:inference_pipeline} demonstrates this process using a representative patient with two lesions---one malignant and one benign---showing how each lesion is independently evaluated through the complete pipeline.

For each detected lesion, a $256 \times 256$ patch is extracted and processed through RGB fusion of the pre-contrast, early post-contrast, and late post-contrast DCE-MRI phases. The SegFormer model generates a binary segmentation mask for each patch, predicting the spatial distribution of malignant tissue. We then compute a malignancy score defined as the ratio of predicted malignant pixels to total lesion pixels ($P_{\text{malignant}}/P_{\text{total}}$), yielding continuous values between 0 and 1. In Figure~\ref{fig:inference_pipeline}, the malignant lesion achieves a score of 0.6, indicating that 60\% of its pixels are predicted as malignant, while the benign lesion scores 0.1, with only 10\% of pixels predicted as malignant.

The classification threshold was optimized through systematic evaluation on the validation set. Comparative analysis of thresholds at 0.3 and 0.5 revealed that the 0.3 threshold achieved superior performance metrics, particularly improving sensitivity for smaller malignant lesions without substantially compromising specificity. Using this optimized threshold, lesions with scores $\geq$0.3 are classified as malignant (output = 1), while those below are classified as benign (output = 0). This pixel-ratio approach provides an interpretable metric that correlates with the spatial extent of malignant features, offering clinicians quantitative insight into the model's decision-making process beyond binary classification.

\begin{table}[ht]
\centering
\caption{Lesion-level Classification and Malignant Segmentation Performance}
\label{tab:classification_performance}
\begin{tabular}{|l|c|c|c|c|c|c|c|c|}
\hline
\textbf{Set} & \textbf{N} & \textbf{TP} & \textbf{TN} & \textbf{FP} & \textbf{FN} & \textbf{Accuracy} & \textbf{AUC} & \textbf{Mean Dice} \\
\hline
Training & 89 & 48 & 14 & 22 & 5 & 0.70 & 0.72 & 0.507 \\
Validation & 15 & 8 & 3 & 2 & 2 & 0.73 & 0.80 & 0.481 \\
Test & 28 & 15 & 8 & 5 & 0 & 0.82 & 0.92 & 0.543 \\
\hline
\end{tabular}
\end{table}

\subsection{Segmentation and Classification Performance}

\begin{figure*}[t]
\centering
\includegraphics[width=\textwidth]{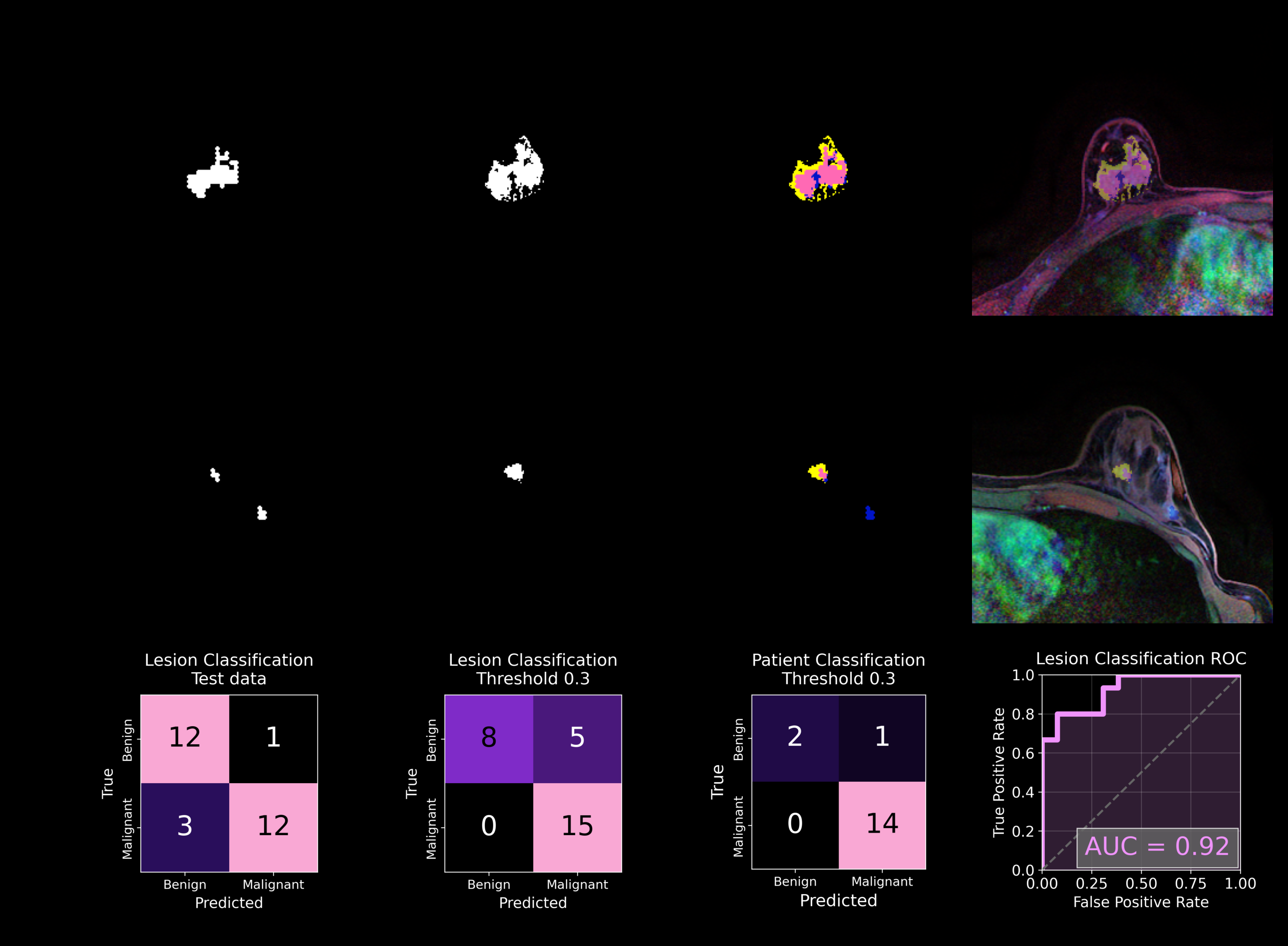}
\caption{Segmentation performance and classification results on test data. Representative examples show (a) malignant lesion with approximately 60\% overlap correctly classified and (b) benign lesion with $<$20\% overlap correctly classified as benign. Columns display ground truth masks, model predictions, overlap visualization (red: true positive, yellow: false positive, blue: false negative pixels), and RGB fusion overlay. Panel (c) presents confusion matrices for lesion-level classification (thresholds 0.5 and 0.3), patient-level classification (threshold 0.3), and ROC curve with AUC = 0.92.}
\label{fig:segmentation_results}
\end{figure*}

The SegFormer-based classification pipeline achieved robust performance on the BreastDCEDL\_AMBL test set, with an AUC of 0.92 for lesion-level classification (Figure~\ref{fig:segmentation_results}c, rightmost panel). Figure~\ref{fig:segmentation_results} demonstrates the model's segmentation capabilities across different lesion types. Malignant lesions (Figure~\ref{fig:segmentation_results}a) exhibited substantial overlap between predicted and ground truth masks (approximately 60\% pixel agreement), while benign lesions (Figure~\ref{fig:segmentation_results}b) showed minimal segmentation activation ($<$20\% overlap), reflecting the model's learned discrimination between tissue types. Results for training, validation, and test sets are summarized in Table~\ref{tab:classification_performance}.

Initial evaluation using a 0.5 threshold yielded 24/28 correct lesion classifications (Figure~\ref{fig:segmentation_results}c, first confusion matrix). However, prioritizing clinical safety necessitated threshold optimization to minimize false negatives. The 0.3 threshold, validated across training and validation sets, achieved zero false negatives while maintaining diagnostic specificity: 8 true negatives, 5 false positives, 0 false negatives, and 15 true positives at the lesion level (Figure~\ref{fig:segmentation_results}c, second confusion matrix).

Patient-level classification, where each patient was classified based on their most suspicious lesion, demonstrated enhanced performance with the optimized threshold: 2 true negatives, 1 false positive, 0 false negatives, and 14 true positives (Figure~\ref{fig:segmentation_results}c, third confusion matrix), achieving 100\% sensitivity for cancer detection. The overlap visualizations (Figure~\ref{fig:segmentation_results}, column 3) reveal the model's segmentation patterns, with correctly identified regions (red), over-segmentation (yellow), and under-segmentation (blue) providing interpretable insights into classification decisions.

\subsection{Technical Limitations}

The current implementation has three primary limitations. First, 2D patch processing sacrifices volumetric context that 3D architectures could leverage for improved morphological assessment. Second, pixel-based metrics ignore physical voxel dimensions, potentially causing inconsistent classifications across different scanning protocols. Third, the model operates on pre-detected lesions rather than providing end-to-end detection and classification. These constraints reflect the trade-off between computational feasibility and the limited scale of publicly available annotated data.

\subsection{Future Directions}

Future work should address current limitations through several key directions. The development of three-dimensional transformer architectures would leverage full volumetric information inherent in MRI data, potentially capturing morphological features lost in our current slice-based approach. Multi-task learning frameworks could enable simultaneous detection, segmentation, and classification of lesions, providing an end-to-end solution for clinical workflows. Integration of multiparametric MRI sequences beyond DCE imaging would enrich the feature space available for classification decisions. Federated learning approaches offer a promising avenue for training models across multiple institutions while preserving data privacy, potentially accessing the scale necessary for robust model development. Finally, prospective validation studies within actual clinical workflows remain essential for assessing real-world performance and impact on patient outcomes. These advances, building upon the foundation established by this benchmark dataset, could accelerate the translation of AI-assisted diagnosis into routine clinical practice.

\subsection{Dataset Contribution}

BreastDCEDL\_AMBL addresses a critical infrastructure gap as the only public DCE-MRI dataset with comprehensive benign and malignant annotations. While containing only 88 patients, it enables reproducible benchmarking---something impossible with the private datasets dominating current literature. The successful training augmentation using ISPY datasets (1,200+ patients, though limited to primary tumors) demonstrates effective transfer learning strategies for overcoming data scarcity.

\subsection{Clinical Translation Potential}

Multi-center validation using standardized voxel normalization is essential before clinical deployment. Integration of clinical variables (age, BRCA status, tumor volume) could improve performance, as demonstrated in prior studies reporting 5--10\% AUC gains. The progression to fully automated detection requires substantially larger annotated datasets, potentially achievable through federated learning approaches that preserve data privacy while enabling large-scale model training.

\section{Conclusion}

We demonstrate that competitive breast lesion classification (AUC 0.92) is achievable using publicly available data and open-source methods, challenging the field's dependence on proprietary resources. By releasing BreastDCEDL\_AMBL and our trained models, we provide essential infrastructure for reproducible research in breast MRI analysis. Future work should focus on volumetric processing, physical dimension normalization, and expanding public datasets to enable robust clinical translation.

\section{Data Availability}

All datasets and code developed in this study are publicly available to ensure reproducibility:

\subsection*{Primary Dataset}
\begin{itemize}
    \item \textbf{BreastDCEDL\_AMBL}: The curated DCE-MRI dataset with benign and malignant lesion annotations (88 patients, 133 lesions)
    \begin{itemize}
        \item Dataset: \url{https://doi.org/10.5281/zenodo.17189101}
        \item Code: \url{https://github.com/naomifridman/BreastDCEDL_AMBL}
    \end{itemize}
\end{itemize}

\subsection*{Training Expansion Datasets}
\begin{itemize}
    \item \textbf{BreastDCEDL\_ISPY1/2}: Additional DCE-MRI datasets used for training augmentation (1,145 patients)
    \begin{itemize}
        \item Dataset: \url{https://doi.org/10.5281/zenodo.15627233}
        \item Code: \url{https://github.com/naomifridman/BreastDCEDL}
    \end{itemize}
\end{itemize}

\subsection*{Source Data}
Original data were obtained from The Cancer Imaging Archive (TCIA):
\begin{itemize}
    \item AMBL collection~\cite{daciels2025ambl_20}: \url{https://www.cancerimagingarchive.net/collection/advanced-mri-breast-lesions}
    \item I-SPY 1~\cite{newitt2016ispy1_33} and I-SPY 2~\cite{li2019ispy2dw_34,wang2022ispy2tcia_35,wang2019ispy2trial_36} trials
\end{itemize}

All materials are released under open licenses and are freely accessible for research, reproducibility, and further development.

\end{document}